%% file: main.tex
\documentclass{article}

\usepackage[preprint]{neurips_2025}

\usepackage[utf8]{inputenc} 
\usepackage[T1]{fontenc}    
\usepackage{hyperref}       
\usepackage{url}            
\usepackage{booktabs}       
\usepackage{amsfonts}       
\usepackage{nicefrac}       
\usepackage{microtype}      
\usepackage{xcolor}         
\usepackage{colortbl}

\usepackage{graphicx} 
\usepackage{wrapfig}
\usepackage{multirow}
 \usepackage{amsmath}

\newcommand {\best}[1]{{\color{red}\textbf{#1}}}
\newcommand {\second}[1]{{\color{blue}\textbf{#1}}}
\newcommand {\improve}[1]{\scriptsize{\color{red}(#1)}}

\def\eg{{\it{e.g.}}}
\def\etal{{\it et al.}}
\def\ie{{\it{i.e.}}}

\definecolor{tabgray}{rgb}{0.95,0.95,0.95}

\title{
Learning Unpaired Image Dehazing with \\
Physics-based Rehazy Generation
}

\author{%
	\textbf{Haoyou Deng $^{1}$}
    ~\textbf{Zhiqiang Li $^{1,2}$}
    ~\textbf{Feng Zhang $^2$}
    ~\textbf{Qingbo Lu $^2$}
    ~\textbf{Zisheng Cao $^2$} \\
    ~\textbf{Yuanjie Shao $^{3}$}
    ~\textbf{Shuhang Gu $^{4}$}
    ~\textbf{Changxin Gao $^{1}$}
    ~\textbf{Nong Sang $^1$}\thanks{Corresponding author.}
    \vspace{1em}
    \\
	$^{1}$ National Key Laboratory of Multispectral Information Intelligent Processing Technology,\\ School of Artificial Intelligence and Automation, Huazhong University of Science and Technology,\\
    $^{2}$ DJI Technology Co., Ltd, ~~
    $^{3}$ School of Electronic Information and Communications,\\ Huazhong University of Science and Technology, \\
    $^{4}$ University of Electronic Science and Technology of China
    \\
    {\tt\small\{haoyoudeng, zhiqiangli, shaoyuanjie, cgao, nsang\}@hust.edu.cn, } \\
    {\tt\small \{fengzhangaia, shuhanggu\}@gmail.com, }
    {\tt\small \{qingbo.lu, zisheng.cao\}@dji.com}
}

\begin{document}

\maketitle

\input{Sec/00_Abstract}    
\input{Sec/01_Intro}
\input{Sec/02_Related}
\input{Sec/03_Method}
\input{Sec/04_Experiment}

\input{Sec/05_Conclusion}

\bibliographystyle{plain}
\bibliography{reference}

\input{Sec/appendix}

\end{document}

%% file: Sec/00_Abstract.tex
\begin{abstract}

Overfitting to synthetic training pairs remains a critical challenge in image dehazing, leading to poor generalization capability to real-world scenarios.
To address this issue, existing approaches utilize unpaired realistic data for training, employing CycleGAN or contrastive learning frameworks.
Despite their progress, these methods often suffer from training instability, resulting in limited dehazing performance.
In this paper, we propose a novel training strategy for unpaired image dehazing, termed Rehazy, to improve both dehazing performance and training stability. This strategy explores the consistency of the underlying clean images across hazy images and utilizes hazy-rehazy pairs for effective learning of real haze characteristics.
To favorably construct hazy-rehazy pairs, we develop a physics-based rehazy generation pipeline, which is theoretically validated to reliably produce high-quality rehazy images.
Additionally, leveraging the rehazy strategy, we introduce a dual-branch framework for dehazing network training, where a clean branch provides a basic dehazing capability in a synthetic manner, and a hazy branch enhances the generalization ability with hazy-rehazy pairs.
Moreover, we design a new dehazing network within these branches to improve the efficiency, which progressively restores clean scenes from coarse to fine.
Extensive experiments on four benchmarks demonstrate the superior performance of our approach, exceeding the previous state-of-the-art methods by 3.58 dB on the SOTS-Indoor dataset and by 1.85 dB on the SOTS-Outdoor dataset in PSNR.
Our code will be publicly available.

\end{abstract}

%% file: Sec/01_Intro.tex
\section{Introduction}
\label{sec:intro}

Haze is a common atmospheric phenomenon caused by the scattering effect in the atmosphere, which corrupts the visibility of outdoor scenes and impairs the performance of downstream vision tasks, such as recognition and detection~\cite{hahner2021fog, li2023domain}. The degradation of haze effect is commonly modeled using the Atmospheric Scattering Model (ASM)~\cite{mccartney1976asm, narasimhan2002asm}:
\begin{equation}\label{eq:asm}
    \mathbf{I}(x) = \mathbf{J}(x)\mathbf{t}(x) + A(1-\mathbf{t}(x)),
\end{equation}
where $\mathbf{I}(x)$ and $\mathbf{J}(x)$ stand for the hazy image and the corresponding clean image at a position $x$, respectively. $A$ is the atmospheric light, and the transmission map $\mathbf{t}(x)=e^{-\beta \mathbf{d}(x)}$ describes the light after scattering, where $\beta$ is the scattering coefficient and $\mathbf{d}(x)$ is the scene depth.

With the rapid advancement of deep learning, numerous dehazing methods~\cite{li2017aod, qin2020ffa, cai2016dehazenet, zheng2023curricular, song2023dehazeformer, qiu2023mb, zhang2024depth, zhang2025beyond, su2025prior} have been developed to restore clean scenes. These methods typically learn a mapping from hazy to clean images in a supervised manner, training on a large amount of synthetic hazy-clean image pairs. Although they achieved impressive performance on synthetic benchmarks, the overfitting issue on the synthetic images often leads to poor generalization capability in real-world conditions~\cite{yang2024d4}.

Since obtaining plenty of genuine hazy-clean pairs in the real world is hard, recent research has increasingly focused on training with unpaired datasets to address real haze removal.
Existing unpaired dehazing approaches generally follow two paradigms: CycleGAN-like methods and CUT-like methods.
Specifically, CycleGAN-like methods~\cite{engin2018cycledehaze, chen2022unpaired, yang2024d4, lan2025exploiting} build  upon the CycleGAN framework~\cite{zhu2017cyclegan} by establishing hazy-clean-hazy and clean-hazy-clean cycles. Through adversarial training, these methods synthesize realistic hazy-clean image pairs, thereby enabling the learning of both the mapping from hazy to clean images and its inverse.
In contrast, CUT-like methods~\cite{chen2022unpaired, wang2024ucl, wang2024odcr, luo2025farewell} adopt contrastive learning~\cite{park2020contrastive} by constructing query-positive-negative pairs sampled from the hazy input, clean input, and dehazed output. Subsequently, a dehazing network is trained by maximizing the mutual information between features of corresponding patches.
Despite these significant advancements, both the adversarial training of CycleGAN-like methods and the contrastive learning of CUT-like methods suffer from training instability, resulting in limited dehazing performance.

To improve both dehazing performance and training stability, we propose a novel training strategy for unpaired image dehazing, termed Rehazy.
The core idea is to explore the clean consistency property, \ie, \textit{a set of hazy images, captured under varying atmospheric scattering conditions but depicting the same scene, is supposed to share the same underlying clean counterpart}. Therefore, we define ``\textit{rehazy images}'' as hazy images that maintain clean consistency with a given hazy image.
Thus, by obtaining hazy-rehazy image pairs and enforcing consistency between their dehazed outputs, the rehazy strategy favorably addresses the absence of real training pairs and enables effective learning of real-world dehazing capabilities.
Additionally, to construct hazy-rehazy pairs, we develop a physics-based rehazy generation pipeline grounded on the ASM model with a solid theoretical proof. Without any learnable parameters, this pipeline reliably generates high-quality rehazy images from a given hazy image.
Furthermore, leveraging this strategy, we introduce a dual-branch framework to train dehazing networks, where a clean branch utilizes a synthetic approach to obtain a basic dehazing capability, and a hazy branch adopts the rehazy strategy to enhance the generalization ability to real hazy scenes.
Within these branches, we also design a multi-scale dehazing network, MS-DehazeNet, which progressively removes haze from coarse to fine, thereby enhancing both dehazing performance and efficiency.
Extensive experimental results on four benchmark datasets demonstrate that the proposed method favorably outperforms existing state-of-the-art approaches.

In conclusion, the main contributions of this work can be summarized into four points:

(1) We propose a novel training strategy for unpaired image dehazing by exploring the clean consistency property. By constructing hazy-rehazy pairs, this strategy facilitates effective learning of real-world haze characteristics.

(2) We develop a physics-based rehazy generation pipeline, which is theoretically validated to reliably produce high-quality rehazy images.

(3) We introduce a dual-branch framework for dehazing network training by leveraging the proposed strategy and design a multi-scale dehazing network to enhance the efficiency of haze removal.

(4) We conduct extensive experiments on four benchmark datasets. Both qualitative and quantitative results demonstrate that our method achieves state-of-the-art performance.

%% file: Sec/02_Related.tex
\vspace{-0.5em}
\section{Related Work} \label{sec:related}

\vspace{-0.25em}
\paragraph{Paired Image Dehazing.}

In real-world scenarios, acquiring a large number of paired hazy and clean images is quite challenging, which limits the application of deep learning in dehazing tasks. To tackle this limitation, Li~\etal~\cite{li2018reside} constructed a large-scale benchmark consisting of various synthetic pairs, enabling the effective training of neural networks. Based on such synthetic data, existing dehazing methods broadly fall into two groups. The first group~\cite{cai2016dehazenet, ren2016single, li2017aod} employs neural networks to estimate the transmission map or atmospheric light from hazy images and adopts the ASM model to restore clean scenes, while the second group~\cite{liu2019griddehazenet, qin2020ffa, qu2019enhanced, zheng2023curricular, qiu2023mb, song2023dehazeformer, zhang2025beyond, su2025prior} directly predicts the clean image with an end-to-end network. Although achieved remarkable advancement, these methods easily overfit to synthetic training data and generalize poorly to real-world hazy scenes~\cite{yang2024d4}.

\vspace{-0.5em}
\paragraph{Unpaired Image Dehazing.}

To overcome the overfitting issue of training on synthetic pairs, several approaches aim to learn the capability of haze removal using unpaired training data. These methods typically follow two paradigms: CycleGAN-like methods and CUT-like methods.
CycleGAN-like methods~\cite{engin2018cycledehaze, chen2022unpaired, yang2024d4, lan2025exploiting} tackle this problem within the image-to-image translation framework, leveraging adversarial loss to generate hazy-clean pairs for network training.
For instance, Yang~\etal~\cite{yang2024d4} decomposed the transmission map into density and depth for hazy for unpaired haze synthesis and removal within a CycleGAN-like framework.
In contrast, CUT-like methods~\cite{chen2022unpaired, wang2024ucl, wang2024odcr, luo2025farewell} employ contrastive learning~\cite{park2020contrastive} to train the dehazing network, which focuses on the construction of query-positive-negative pairs.
For example, Wang~\etal~\cite{wang2024odcr} introduced Orthogonal-MLPs to decouple features into haze-related and haze-unrelated components, maximizing the mutual information between the corresponding components of query and positive samples.
Despite this progress, the training instability in both adversarial loss and contrastive learning limits their dehazing performance.
To tackle this limitation, we propose a novel training strategy by exploring clean consistency and constructing hazy-rehazy pairs, enabling effective training for dehazing networks.

%% file: Sec/03_Method.tex
\section{Methodology}

\begin{figure}
    \centering
    \includegraphics[width=1.0\linewidth]{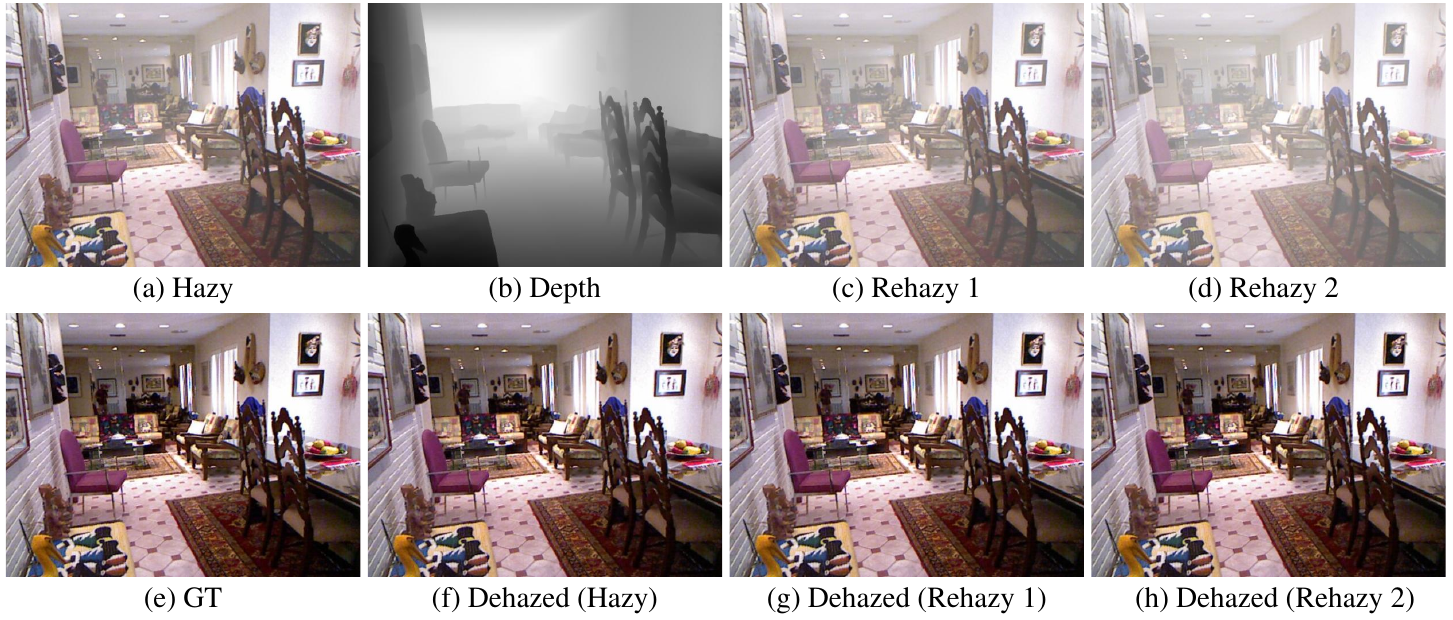}
    \vspace{-1.5em}
    \caption{Visualization of rehazy generation. Given a hazy image (a), we estimate its depth (b) and subsequently adopt Eq.~\ref{eq:rehazy} to generate rehazy images (c)(d). One can see that their corresponding dehazed results (f)(g)(h) exhibit high similarity with the ground truth (e).}
    \vspace{-1em}
    \label{fig:rehazy_vis}
\end{figure}

\subsection{Motivation} \label{sec:motivation}

Typically, supervised approaches train a dehazing network $g_\theta$ (parameterized by $\theta$) with plenty of hazy images $\mathbf{I}\in \mathbb{R}^{H\times W\times 3}$ and their corresponding clean images $\mathbf{J}\in \mathbb{R}^{H\times W\times 3}$, by optimizing the objective as:
\begin{equation} \label{eq:pair_obj}
\mathop{\rm argmin~}\limits_{\theta} \mathbb{E}_{(\mathbf{I},\mathbf{J})}\{\mathcal{L}(g_{\theta}(\mathbf{I}),\mathbf{J})\},
\end{equation}
where $\mathcal{L}$ denotes an empirical loss function.
However, since collecting plenty of suitable real-world pairs $\{(\mathbf{I}, \mathbf{J})\}$ is challenging, optimizing Eq.~\ref{eq:pair_obj} lacks feasibility for real conditions.
To eliminate this issue, we propose the rehazy strategy driven by the clean consistency, which effectively optimizes Eq.~\ref{eq:pair_obj} using unpaired realistic clean and hazy images.
Specifically, considering that a set of hazy images $\{\mathbf{I}_i\}(i=0,1,...,n)$ maintain the clean consistency and $\mathbf{J}_0$ is their shared clean counterpart, the optimization objective in Eq.~\ref{eq:pair_obj} becomes equivalent to
\begin{equation} \label{eq:rehazy_obj}
    \mathop{\rm argmin}\limits_{\theta}~\mathbb{E}_{\{\mathbf{J}_0\}}~[~
    \mathop{\underline{
    \mathbb{E}_{(i\neq0)}\{\mathcal{L}(g_{\theta}(\mathbf{I}_i),g_{\theta}(\mathbf{I}_0))\}
    }}\limits_{\mathcal{L}_{const}}
    +\mathop{\underline{
    \mathcal{L}(g_{\theta}(\mathbf{I}_0),\mathbf{J}_0)
    }}\limits_{\mathcal{L}_{dehazing}}
    ~],
\end{equation}
In Eq.~\ref{eq:rehazy_obj}, the first term $\mathcal{L}_{const}$ enforces the clean consistency among the dehazed results of $\{\mathbf{I}_i\}$, while the second term $\mathcal{L}_{dehazing}$ aligns the dehazed output of $\mathbf{I}_0$ with the clean ground truth $\mathbf{J}_0$ to allow effective learning of dehazing capability.

We observe that in the absence of paired data, optimizing Eq.~\ref{eq:rehazy_obj} is considerably more feasible than optimizing Eq.~\ref{eq:pair_obj}.
Specifically, in Eq.~\ref{eq:rehazy_obj}, $\mathcal{L}_{const}$ only depends on hazy images, thereby eliminating the requirement for corresponding clean counterparts.
Additionally, though $\mathcal{L}_{dehazing}$ is defined over the unreachable pair $(\mathbf{I}_0, \mathbf{J}_0)$, we draw inspiration from~\cite{fang2024real, wang2025learning} and introduce two strategies that serve as a proxy for this supervision in the unpaired setting.
First, we employ the ASM model to synthesize hazy images from clean images, allowing the network to acquire a basic dehazing capability through supervised training. Second, we exploit the invariant properties shared between hazy and clean images to enhance real-world generalization, as elaborated in Sec.~\ref{sec:framework}.
Consequently, Eq.~\ref{eq:rehazy_obj} can enable effective training of a well-performing dehazing network using unpaired data.
Nevertheless, a critical challenge remains for optimizing Eq.~\ref{eq:rehazy_obj}: how to reliably construct the hazy image set $\{\mathbf{I}_i\}$ with clean consistency?

\subsection{Physics-based Rehazy Generation} \label{sec:rehazy}

Given a real hazy image $\mathbf{I}_0$, an ideal $\{\mathbf{I}_i\}$ can be formed by combining $\mathbf{I}_0$ itself along with its corresponding rehazy images $\{\mathbf{I}_r\}$ that share the same underlying clean content.
To reliably construct $\{\mathbf{I}_i\}$, we introduce a physics-based rehazy generation pipeline to produce the favorable rehazy image $\mathbf{I}_r$.
Specifically, based on the ASM model and Eq.~\ref{eq:asm}, $\mathbf{I}_0$ and $\mathbf{I}_r$ can be formulated as:
\begin{equation} \label{sec:asm_I0}
    \mathbf{I}_0=\mathbf{J}_0 \mathbf{t}_0+A_0 (1-\mathbf{t}_0),~\mathbf{I}_r=\mathbf{J}_r\mathbf{t}_r+A_r(1-\mathbf{t}_r).
\end{equation}
Here, $\mathbf{J}_0$, $\mathbf{t}_0=e^{-\beta_0 \mathbf{d}_0} \in \mathbb{R}^{H\times W\times 1} $, and $A_0 \in \mathbb{R}^{3}$ denote the clean counterpart, transmission map, and airlight of $\mathbf{I}_0$, respectively, where $\beta_0 \in \mathbb{R}^{1}$ is the scattering coefficient and $\mathbf{d}_0 \in \mathbb{R}^{H\times W\times 1}$ is the depth of $\mathbf{J}_0$. Similarly, $\mathbf{J}_r$, $\mathbf{t_r}=e^{-\beta_r \mathbf{d}_r}$, and $A_r$ correspond to $\mathbf{I}_r$.
Due to the clean consistency assumption, we have $\mathbf{J}_0=\mathbf{J}_r$ and $\mathbf{d}_0=\mathbf{d}_r$.
Letting $A_r=A_0+\Delta A$, $\beta_r=\beta_0+\Delta \beta$, and $\mathbf{t}_r=\mathbf{t}_0\Delta \mathbf{t}$, we derive $\Delta \mathbf{t}=e^{-\Delta \beta \mathbf{d}_0}$ and can thus rewrite Eq.~\ref{sec:asm_I0} as:
\begin{equation} \label{eq:Ir2I0}
\begin{aligned}
    \mathbf{I}_r&=\mathbf{J}_0\mathbf{t}_0\Delta \mathbf{t}+(A_0+\Delta A)(1-\mathbf{t}_0\Delta \mathbf{t}) \\
    &=\Delta \mathbf{t}(\mathbf{J}_0 \mathbf{t}_0 +A_0(1-\mathbf{t}_0)) + A_0 (1-\Delta \mathbf{t}) +\Delta A(1-\mathbf{t}_0\Delta \mathbf{t}) \\
    &=\Delta \mathbf{t} \mathbf{I}_0 + A_0(1-\Delta \mathbf{t}) +\Delta A(1-\mathbf{t}_0\Delta \mathbf{t})
\end{aligned}.
\end{equation}
Furthermore, for simplification, we assume a shared airlight between $\mathbf{I}_0$ and $\mathbf{I}_r$, that is, $A_0=A_r$, \ie, $\Delta A=0$. This assumption indicates that the generated images constitute a subset of the full rehazy images distribution, and we discover it still beneficial in this way.
With this assumption, the Eq.~\ref{eq:Ir2I0} is equivalent to
\begin{equation} \label{eq:rehazy}
    \mathbf{I}_r = \Delta \mathbf{t} \mathbf{I}_0 + A_0(1-\Delta \mathbf{t}),~\Delta \mathbf{t}=e^{-\Delta \beta \mathbf{d}_0}.
\end{equation}
This suggests that the generation of $\mathbf{I}_r$ from $\mathbf{I}_0$ can be determined by only three parameters: $A_0$, $\Delta \beta$, and $\mathbf{d}_0$. In practice, we employ three approaches to estimate these parameters.
First, we utilize the DCP~\cite{he2010dcp} method to estimate $A_0$ from $\mathbf{I}_0$.
Second, following previous studies~\cite{li2018reside, yang2024d4}, $\Delta \beta$ is randomly sampled from a pre-defined distribution ${\rm U}(\Delta \beta_{min}, \Delta \beta_{max})$, where ${\rm U}(\cdot)$ denotes a uniform distribution.
Third, as ~\cite{ye2025prompthaze} demonstrates that the Depth Anything model~\cite{yang2024depth, yang2024depthv2} has the ability to produce stable and reliable depth estimations under various hazy conditions, a pre-trained Depth Anything model~\cite{yang2024depthv2} is employed to predict the depth of $\mathbf{I}_0$ as an approximation of $\mathbf{d}_0$. 
Using these approaches, as shown in Fig.~\ref{fig:rehazy_vis}, Eq.~\ref{eq:rehazy} can be effectively applied to generate favorable rehazy images from real hazy inputs in a physics-driven manner without any learnable parameters.

\begin{figure}
    \centering
    \includegraphics[width=1.0\linewidth]{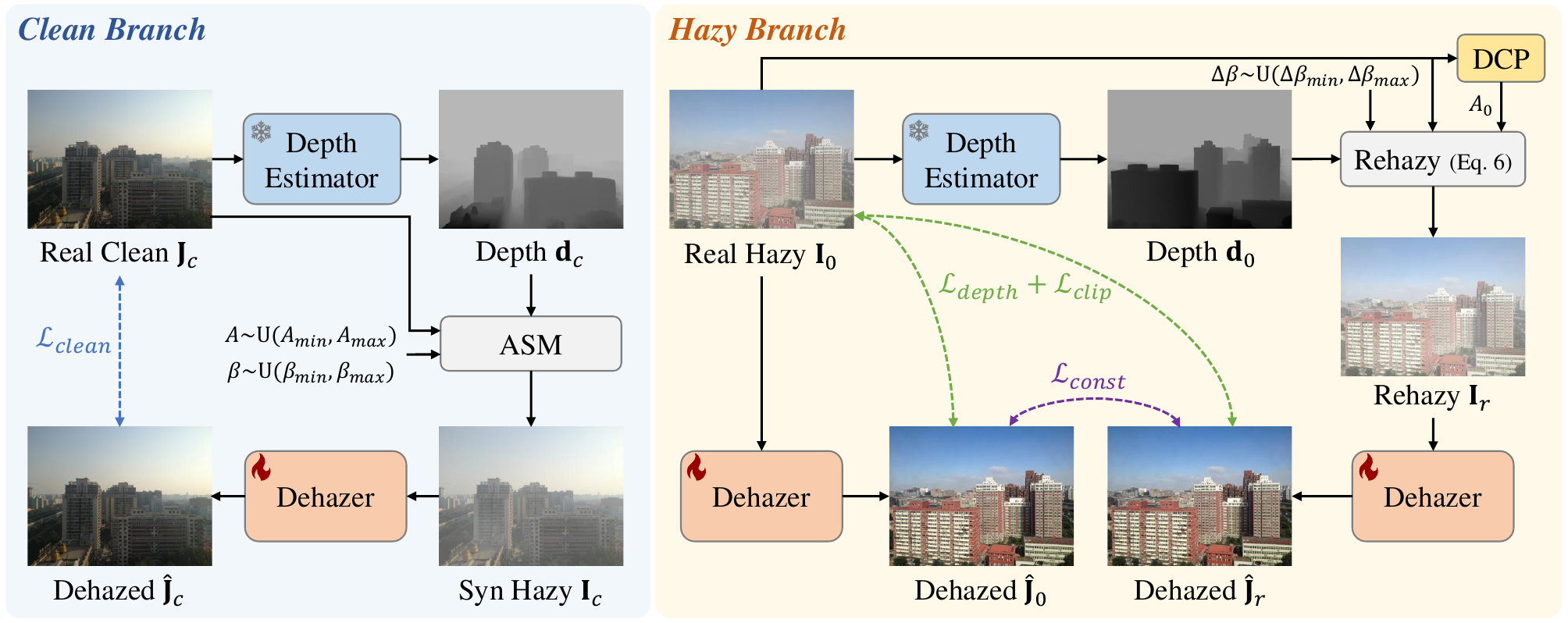}
    \vspace{-1.6em}
    \caption{Overview of our dual-branch framework: a clean branch learns fundamental dehazing capabilities in a synthetic manner, while a hazy branch captures real-world haze characteristics from hazy-rehazy pairs using the rehazy strategy.}
    \label{fig:framework}
    \vspace{-1em}
\end{figure}

\vspace{-0.25em}
\subsection{Overall Framework} \label{sec:framework}
\vspace{-0.25em}

Leveraging the rehazy strategy, we develop a dual-branch framework for dehazing network training, as illustrated in Fig.~\ref{fig:framework}. The framework comprises a fixed depth estimator, implemented using a pre-trained Depth Anything model~\cite{yang2024depthv2}, and a dehazer for haze removal based on MS-DehazeNet (detailed in Sec~\ref{sec:dehazer}), with both networks sharing weights across the two branches.
As discussed in Sec.~\ref{sec:motivation}, we adopt a synthetic manner to learn a basic dehazing capability using clean images $\mathbf{J}_c$ in the clean branch, and employ the rehazy strategy to capture real-world haze features from unpaired hazy images $\mathbf{I}_0$ in the hazy branch.

\vspace{-0.5em}
\paragraph{Clean branch.}
The clean branch adopts the ASM model to produce hazy images from clean images and trains the dehazer in a supervised manner. Specifically, feeding a clean image $\mathbf{J}_c$, we initially use the depth estimator to predict its depth $\mathbf{d}_c$. Besides, we randomly sample an atmospheric light $A\sim {\rm U}(A_{min}, A_{max})$ and a scattering coefficient $\beta \sim {\rm U}(\beta_{min}, \beta_{max})$. Using $\{\mathbf{J}_c, \mathbf{d}_c, A,\beta\}$, a synthetic hazy image $\mathbf{I}_c$ can be generated via Eq.~\ref{eq:asm}. By this means, plenty of synthetic hazy images with their clean counterparts can be produced to train the dehazer using a supervised loss:
\begin{equation} \label{eq:clean_loss}
    \mathcal{L}_{clean} = \Vert \hat{\mathbf{J}}_c-\mathbf{J}_c \Vert_1,
\end{equation}
where $\hat{\mathbf{J}}_c$ is the dehazed result of $\mathbf{I}_c$ using the dehazer and $\Vert \cdot \Vert_1$ denotes L1 loss.
Accordingly, this supervised process enables the dehazer to learn favorable dehazing capabilities on synthetic data.

\vspace{-0.5em}
\paragraph{Hazy branch.}
Given a hazy image $\mathbf{I}_0$ as input, the hazy branch generates rehazy images and learns with clean consistency via the rehazy strategy.
Specifically, we first utilize the depth estimator to predict the depth map $\mathbf{d}_0$ and employ DCP~\cite{he2010dcp} method to estimate the airlight $A_0$. With a randomly sampled $\Delta \beta \sim {\rm U}(\Delta \beta_{min}, \Delta \beta_{max})$, a rehazy image $\mathbf{I}_r$ is generated via Eq.~\ref{eq:rehazy}. Subsequently, we adopt the dehazer to remove haze within $\mathbf{I}_0$ and $\mathbf{I}_r$, yielding dehazed outputs $\hat{\mathbf{J}}_0$ and $\hat{\mathbf{J}}_r$, respectively.
With these generated hazy-rehazy pairs, we employ three loss functions to optimize the objective described in Eq.~\ref{eq:rehazy_obj}.
Specifically, to minimize first term of Eq.~\ref{eq:rehazy_obj}, an consistency loss is designed as:
\begin{equation} \label{eq:const_loss}
    \mathcal{L}_{const} = \Vert \hat{\mathbf{J}}_0 - \hat{\mathbf{J}}_r \Vert_1,
\end{equation}
which enforces consistency between the dehazed outputs $\hat{\mathbf{J}}_0$ and $\hat{\mathbf{J}}_r$.
Furthermore, for the second term of Eq.~\ref{eq:rehazy_obj}, we further exploit two invariant properties shared between hazy and dehazed images to improve generalization for real scenes. One is that due to the robust capability of the Depth Anything model, the dehazed results should have a similar depth to the hazy input~\cite{ye2025prompthaze}, described as:
\begin{equation}
    \mathcal{L}_{depth} = \Vert {\rm D}(\hat{\mathbf{J}}_0) - {\rm D}(\mathbf{I}_0) \Vert_1 + \Vert {\rm D}(\hat{\mathbf{J}}_r) - {\rm D}(\mathbf{I}_0) \Vert_1,
\end{equation}
where ${\rm D}(\cdot)$ represents a pre-trained Depth Anything model.
Besides, another invariance lies in the consistency of semantic information during the dehazing process. Hence, we leverage the strong semantic representation capability of the CLIP~\cite{radford2021clip} model to extract semantic features from both hazy and dehazed images, and a KL divergence loss is applied to regularize the extracted features:
\begin{equation}
    \mathcal{L}_{clip}={\rm KL}({\rm CLIP}(\hat{\mathbf{J}}_0),{\rm CLIP}(\mathbf{I}_0)) + {\rm KL}({\rm CLIP}(\hat{\mathbf{J}}_r),{\rm CLIP}(\mathbf{I}_0)),
\end{equation}
where ${\rm KL}(\cdot)$ denotes Kullback–Leibler divergence and ${\rm CLIP}(\cdot)$ is the CLIP model.
With these designs, the invariances facilitate effective learning of real haze and contribute to a more stable training process.
To summarize, the training objective of the hazy branch is combined as follows:
\begin{equation}
    \mathcal{L}_{hazy} = \mathcal{L}_{const} + \lambda_{depth}\mathcal{L}_{depth} + \lambda_{clip}\mathcal{L}_{clip},
\end{equation}
where $\lambda_{depth}$ and $\lambda_{clip}$ are the balancing hyper-parameters, which are set to $0.01$ and $0.01$ in our experiment, respectively.
Consequently, the hazy branch enables effective learning from real hazy images, enhancing the dehazing capability in the wild.
To this end, in the training process, we first train the clean branch for $3\times10^5$ iterations with the loss $\mathcal{L}_{clean}$, and then train both clean and hazy branches for $1\times10^5$ iterations with the total loss $\mathcal{L}_{total} = \mathcal{L}_{clean} + \mathcal{L}_{hazy}$.

\vspace{-0.5em}
\subsection{Dehazing Network} \label{sec:dehazer}
\vspace{-0.25em}

\begin{wrapfigure}{r}{0.63\textwidth}
\vspace{-0.55em}
    \begin{center}
        \includegraphics[width=0.6\textwidth]{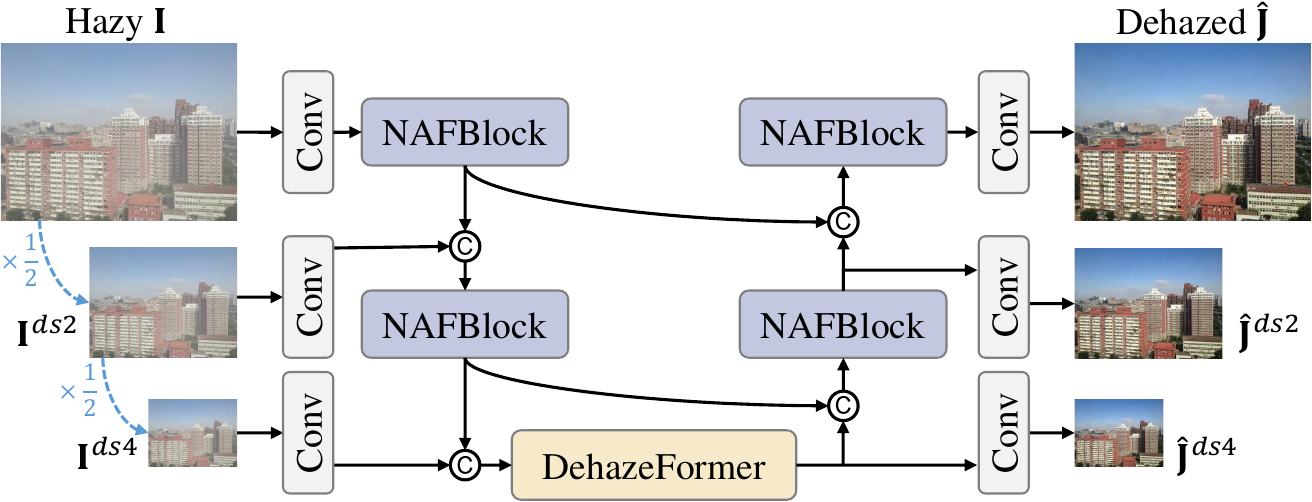}
    \end{center}
    \vspace{-0.5em}
    \caption{Architecture of the proposed MS-DehazeNet.}
    \vspace{-0.75em}
    \label{fig:dehazer}
\end{wrapfigure}

As shown in Fig.~\ref{fig:framework}, a dehazer is employed for haze removal, which can be implemented using any image dehazing network, such as the state-of-the-art DehazeFormer~\cite{song2023dehazeformer}.
Despite its strong performance, DehazeFormer exhibits two limitations: (1) its transformer-based architecture incurs substantial computational overhead; (2) its patch cropping operation disrupts structural details, particularly along object boundaries.
To tackle these issues, we design a three-scale U-Net-like dehazing network, termed MS-DehazeNet, enhancing both dehazing performance and computational efficiency.
As depicted in Fig.~\ref{fig:dehazer}, given that haze primarily impairs the low-frequency components~\cite{guo2023enhanced}, the lowest scale incorporates a superior DehazeFormer to ensure a high-quality restoration. Additionally, to mitigate boundary artifacts and reduce complexity, instead of transformer-based modules, the first and second scales adopt the block of NAFNet~\cite{chen2022nafnet} (\ie~NAFBlock) as the encoder and decoder, which is a lightweight yet effective non-transformer block tailored for image restoration tasks.

Furthermore, inspired by~\cite{cho2021rethinking,cui2023fsnet}, MS-DehazeNet adopts a multi-input, multi-output mechanism. Specifically, an input hazy image $\mathbf{I}$ is initially downsampled by factors of $1/2$ and $1/4$ to $\mathbf{I}^{ds2}$ and $\mathbf{I}^{ds4}$, respectively, followed by convolutional layers extracting features.
These features are then fed into the corresponding scales: directly at the first scale, and concatenated with the outputs of the higher scale at other scales.
After encoding and decoding in a U-shape manner~\cite{ronneberger2015unet}, the outputs of decoders are further processed by additional convolutional layers to reconstruct the dehazed outputs as $\hat{\mathbf{J}}$, $\hat{\mathbf{J}}^{ds2}$, and $\hat{\mathbf{J}}^{ds4}$.
In addition, to ease training difficulty, we constrain the outputs of each scale using their corresponding ground truths. Hence, $\mathcal{L}_{clean}$ in Eq.~\ref{eq:clean_loss} and $\mathcal{L}_{const}$ in Eq.~\ref{eq:const_loss} are updated as:
\begin{equation} \label{eq:ms_loss}
\begin{aligned}
\mathcal{L}_{clean} &= \Vert \hat{\mathbf{J}}_c-\mathbf{J}_c \Vert_1 + \lambda_{1/2} \Vert \hat{\mathbf{J}}_c^{ds2}-\mathbf{J}_c^{ds2} \Vert_1 + \lambda_{1/4} \Vert \hat{\mathbf{J}}_c^{ds4}-\mathbf{J}_c^{ds4} \Vert_1 \\
\mathcal{L}_{const} &= \Vert \hat{\mathbf{J}}_0-\hat{\mathbf{J}}_r \Vert_1 + \lambda_{1/2} \Vert \hat{\mathbf{J}}_0^{ds2}-\hat{\mathbf{J}}_r^{ds2} \Vert_1 + \lambda_{1/4} \Vert \hat{\mathbf{J}}_0^{ds4}-\hat{\mathbf{J}}_r^{ds4} \Vert_1 
\end{aligned},
\end{equation} 
where $\lambda_{1/2}$ and $\lambda_{1/4}$ are the balancing hyper-parameters and set to $0.5$ and $0.25$ in our experiment, respectively. In this way, the multi-scale losses guide the network to progressively remove haze from coarse to fine, thereby enhancing its dehazing capability.

%% file: Sec/04_Experiment.tex
\vspace{-0.5em}
\section{Experiments} \label{sec:experiment}
\vspace{-0.25em}

\begin{table*}
    \begin{center}
        \caption{
            Quantitative comparison on SOTS-Indoor, SOTS-Outdoor, and I-HAZE datasets. \best{Red} and \second{blue} indicate the best and the second best performance. ``-'' means that results are not available.
        }
        \vspace{0.5em}
        \centering
        \setlength{\tabcolsep}{1.15pt}
        \scalebox{0.9}{
            \begin{tabular}{l | c c | c c c | c c c | c c c}
                \toprule[0.15em]
                \multirow{2}{*}{Methods} &Params &MACs &\multicolumn{3}{c|}{SOTS-Indoor} &\multicolumn{3}{c|}{SOTS-Outdoor} &\multicolumn{3}{c}{I-HAZE} \\
                & (M) & (G) & PSNR$\textcolor{black}{\uparrow}$ & SSIM$\textcolor{black}{\uparrow}$ & CIEDE$\textcolor{black}{\downarrow}$ & PSNR$\textcolor{black}{\uparrow}$ & SSIM$\textcolor{black}{\uparrow}$ & CIEDE$\textcolor{black}{\downarrow}$ & PSNR$\textcolor{black}{\uparrow}$ & SSIM$\textcolor{black}{\uparrow}$ & CIEDE$\textcolor{black}{\downarrow}$ \\
                \midrule[0.15em]
                DCP~\cite{he2010dcp} & - & - & 13.10 & 0.6990 & 7.404 & 20.15 & 0.9190 & 7.613 & 13.10 & 0.6990 & 19.040 \\
                CycleGAN~\cite{zhu2017cyclegan} & 11.38 & 56.98 & 21.34 & 0.8980 & 7.000 & 20.55 & 0.8560 & 9.298 & 15.29 & 0.7560 & 19.500 \\
                CycleDehaze~\cite{engin2018cycledehaze} & 11.38 & 49.16 & 20.11 & 0.8540 & 8.761 & 21.31 & 0.8990 & 9.481 & 14.69 & 0.7510 & 19.050 \\
                YOLY~\cite{li2021yoly} & 32.00 & - & 15.84 & 0.8190 & 12.370 & 14.75 & 0.8570 & 15.850 & 15.52 & 0.7090 & 14.490 \\
                RefineDNet~\cite{zhao2021refinednet} & 65.80 & 75.41 & 24.36 & 0.9390 & 4.305 & 19.84 & 0.8530 & 8.481 & 13.60 & 0.6600 & 17.080 \\
                D4+~\cite{yang2024d4}  & \second{10.70} & \best{2.25} & 25.79 & 0.9370 & \second{3.510} & \second{26.30} & \second{0.9600} & \second{4.061} &\second{16.17} & \second{0.8160} & \second{13.550} \\
                UCL-Dehaze~\cite{wang2024ucl} & 14.12 & 78.80 & 20.05 & 0.8355 & 8.049 & 18.28 & 0.6843 & 11.473 & 13.67 & 0.5551 & 18.506 \\
                ODCR~\cite{wang2024odcr} & 11.38 & - & \second{26.32} & \second{0.9450} & - & 26.16 & \second{0.9600} & - & - & - & - \\
                \midrule
                \midrule
                \rowcolor{tabgray}
                \textbf{Ours} & \best{1.57} & \second{11.09} & \best{29.90} & \best{0.9709} & \best{2.299} & \best{28.15} & \best{0.9610} & \best{3.823} & \best{16.93}& \best{0.8185} & \best{12.020} \\
              
                \bottomrule[0.15em]
        \end{tabular}}
        \vspace{-1.25em}
        \label{table:results}
    \end{center}
\end{table*}

\begin{table*}
    \begin{center}
        \caption{Quantitative comparison on the RTTS dataset. \best{Red} and \second{blue} indicate the best and the second best performance. ``Det. mAP'' is the mean Average Precision results for the detection task.}
        \vspace{0.5em}
        \centering
        \setlength{\tabcolsep}{1.75pt}
        \scalebox{0.9}{
            \begin{tabular}{l | c c c c c c || >{\columncolor{tabgray}} c}
                \toprule[0.15em]
                Methods & DCP~\cite{he2010dcp} & CycleDehaze~\cite{engin2018cycledehaze} & YOLY~\cite{li2021yoly} & RefineDNet~\cite{zhao2021refinednet} & D4+~\cite{yang2024d4} & UCL-Dehaze~\cite{wang2024ucl} & \textbf{Ours} \\
                \midrule[0.15em]
                FADE$\textcolor{black}{\downarrow}$ & 2.484 & 1.118 & 1.044 &  0.961 & \second{0.955} & 1.060 & \best{0.891} \\
                BRISQUE$\textcolor{black}{\downarrow}$ & 36.642 & 30.473 & 32.120 & 20.447 & 33.366 & \second{20.000} & \best{13.762} \\
                NIMA$\textcolor{black}{\uparrow}$ & 4.483 & \second{4.822} & 4.801 & 4.743 & 4.801 & 4.654 & \best{4.885} \\

                \midrule
                Det. mAP$\textcolor{black}{\uparrow}$ & 0.605 & 0.010 & 0.515 & 0.610 & \second{0.613} & 0.585 & \best{0.616} \\
              
                \bottomrule[0.15em]
        \end{tabular}}
        \label{table:rtts_results}
    \end{center}
    \vspace{-1em}
\end{table*}

\subsection{Experimental Setup} \label{sec:exp_set}

\vspace{-0.25em}
\paragraph{Datasets.}
We train and evaluate our method using RESIDE~\cite{li2018reside} dataset, RESIDE-unpaired~\cite{zhao2021refinednet} and I-HAZE~\cite{ancuti2018ihaze} dataset.
RESIDE is a widely used benchmark for image dehazing, which contains: (1) ITS/OTS, containing 13990/313950 synthetic indoor/outdoor hazy-clean pairs; (2) SOTS-indoor/outdoor, with 500/500 synthetic indoor/outdoor hazy-clean pairs; (3) RTTS, including 4322 real hazy images without ground-truths but with detection annotations. Additionally, RESIDE-unpaired dataset consists of 3577 real outdoor clean images and 2903 hazy images with higher quality selected from RESIDE. Moreover, I-HAZE dataset comprises 35 real hazy/clean indoor pairs.

\vspace{-0.5em}
\paragraph{Competitors \& Metrics.}
We compare our method with several state-of-the-art unpaired dehazing methods, including DCP~\cite{he2010dcp}, CycleGAN~\cite{zhu2017cyclegan}, CycleDehaze~\cite{engin2018cycledehaze}, YOLY~\cite{li2021yoly}, RefineDNet~\cite{zhao2021refinednet}, D4+~\cite{yang2024d4}, UCL-Dehaze~\cite{wang2024ucl}, and ODCR~\cite{wang2024odcr}.
During test, we employ PSNR, SSIM~\cite{wang2004ssim}, and CIEDE2000 (CIEDE for short)~\cite{sharma2005ciede2000} as objective metrics. As the RTTS dataset lacks ground truth, we adopt the non-reference metrics for evaluation: FADE~\cite{choi2015fade}, BRISQUE~\cite{mittal2012brisque}, and NIMA~\cite{talebi2018nima}.
Note that higher PSNR/SSIM/NIMA and lower CIEDE/FADE/BRISQUE indicate better performance.
In addition, to evaluate efficiency, we provide the number of parameters and Multiply–Accumulate Operations (MACs) for $256 \times 256$ input resolution.

\vspace{-0.5em}
\paragraph{Implementation details.}
Following the evaluation strategy of ~\cite{yang2024d4, wang2024odcr}, we train our method and all the competitive methods on the ITS dataset without the paired information, evaluate them on SOTS-indoor/outdoor and I-HAZE datasets, and also train on RESIDE-unpaired dataset for evaluation on RTTS dataset.
To optimize the network, we employ the Adam optimizer~\cite{kingma2014adam} with $\beta_1=0.9$, $\beta_2=0.999$, and initial learning rate $l_r=2\times 10^{-4}$. Additionally, we use a batch size of $2$ with the training size of $256\times256$.
For indoor scenes, the distribution of $A\sim {\rm U}(A_{min}, A_{max})$, $\beta \sim {\rm U}(\beta_{min}, \beta_{max})$, and $\Delta \beta \sim {\rm U}(\Delta \beta_{min}, \Delta \beta_{max})$ is set to ${\rm U}(0.7, 1.0)$, ${\rm U}(0.6, 1.8)$, and ${\rm U}(0.1,0.5)$, respectively. To better adapt to outdoor environments, these values come to ${\rm U}(0.8, 1.0)$, ${\rm U}(0.01, 0.3)$, and ${\rm U}(0.01,0.15)$ for outdoor scenes.
The implementation is conducted on the Pytorch~\cite{paszke2017pytorch} framework with NVIDIA Tesla V100 32GB GPUs.

\begin{figure}
    \centering
    \includegraphics[width=1.0\linewidth]{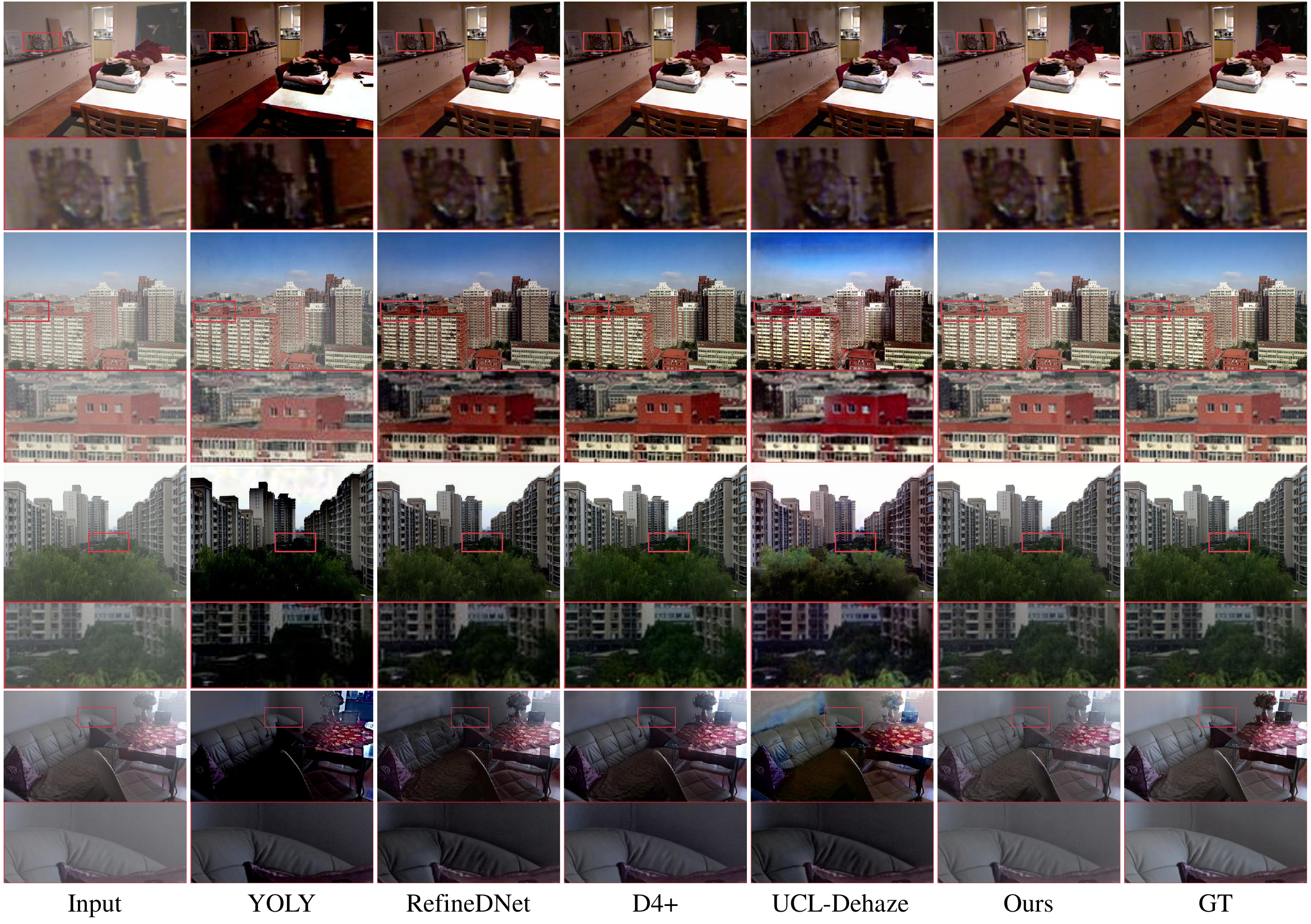}
    \vspace{-1.5em}
    \caption{Visual comparison on samples from SOTS-indoor, SOTS-outdoor and I-HAZE datasets. Best viewed in color and by zooming in.}
    \vspace{-1em}
    \label{fig:results_its_vis}
\end{figure}

\vspace{-0.25em}
\subsection{Quantitative Comparison}
\vspace{-0.25em}

The quantitative comparisons on SOTS-Indoor, SOTS-Outdoor, and I-HAZE datasets are reported in Tab.~\ref{table:results}. One can observe that our method consistently outperforms all competing approaches across all benchmarks, highlighting the effectiveness of the proposed rehazy strategy and framework.
Notably, on SOTS-Indoor dataset, our approach surpasses the second-best method, ODCR~\cite{wang2024odcr}, by a significant margin of 3.58 dB in PSNR, 0.0259 in SSIM, and 1.211 in CIEDE. Additionally, on SOTS-Outdoor and I-HAZE datasets,  it also achieves at least 1.85 dB and 0.76 dB improvements in PSNR, indicating strong generalization ability.
When it comes to real scenes on RTTS dataset, as shown in Tab.~\ref{table:rtts_results}, our method also delivers superior performance. Specifically, it achieves improvements of at least 0.064 in FADE and 6.238 in BRISQUE, demonstrating its robustness in real-world scenarios.
Moreover, thanks to the hierarchical multi-scale architecture of MS-DehazeNet, our method maintains a compact model size (1.57 M) and low computational demands (11.09 G).
Therefore, the achieved superior performance demonstrates the effectiveness and robustness of our methods for unpaired dehazing in both synthetic and real-world scenarios.

\vspace{-0.25em}
\subsection{Qualitative Comparison}

\begin{figure}
    \centering
    \includegraphics[width=1.0\linewidth]{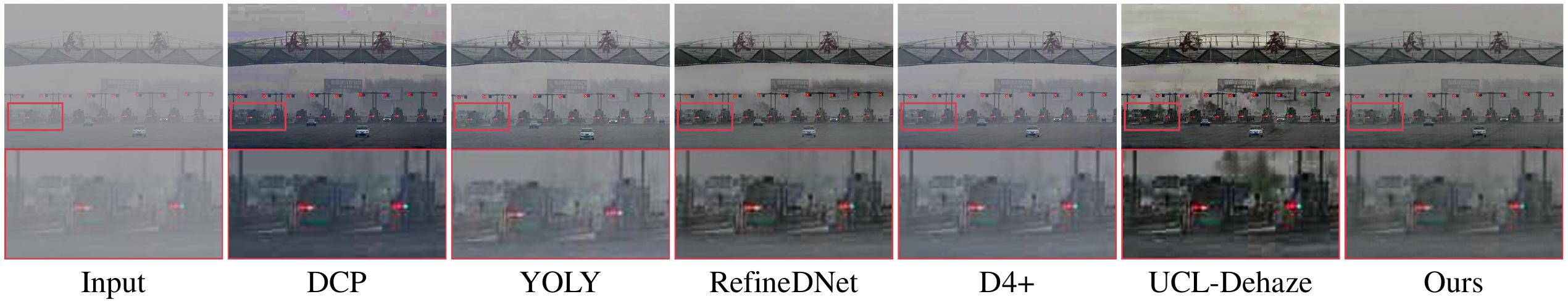}
    \vspace{-1.5em}
    \caption{Visual comparison on samples from RTTS dataset. Best viewed in color and by zooming in.}
    \vspace{-0.75em}
    \label{fig:results_rtts_vis}
\end{figure}

To evaluate our proposed network intuitively, we visually compare dehazed images on the four benchmarks, as shown in Fig.~\ref{fig:results_its_vis} and Fig.~\ref{fig:results_rtts_vis}.
These visual results show that the proposed approach excels in eliminating haze and restoring natural color tones, consistently yielding superior visual quality.
For instance, in the first example of Fig.~\ref{fig:results_its_vis}, while other methods struggle to estimate haze density and consequently disrupt background details, our method accurately identifies haze patterns and produces more natural outputs. Similar advantages are observed across other examples.
Additionally, on the real hazy scene in the last row, our method effectively resolves the haze and yields natural results.
Moreover, on real-world images from the RTTS dataset (Fig.~\ref{fig:results_rtts_vis}), one can see that YOLY and D4+ cannot remove haze clearly, RefineDNet exhibits over-dehazing, and both DCP and UCL-Dehaze introduce artifacts in the sky or at object boundaries. In contrast, our method is able to remove haze accurately and generate natural outputs.
Hence, our method can favorably remove haze and output visually pleasant results, which further validates its superiority.

\begin{table*}
    \centering
    
    \begin{minipage}[c]{0.6\textwidth}
    \caption{Ablation study of rehazy strategy.}
    \vspace{0.5em}
    \centering
    \setlength{\tabcolsep}{3pt}
    \scalebox{0.9}{
    \begin{tabular}{l | c c | c c }
        \toprule[0.15em]
            \multirow{2}{*}{Dehazer} &\multicolumn{2}{c|}{w/o Rehazy} & \multicolumn{2}{c}{w/ Rehazy}  \\
             & PSNR$\textcolor{black}{\uparrow}$ & SSIM$\textcolor{black}{\uparrow}$ & PSNR$\textcolor{black}{\uparrow}$ & SSIM$\textcolor{black}{\uparrow}$  \\
            \midrule[0.15em]
            U-Net~\cite{ronneberger2015unet} & 26.06 & 0.9551 & 27.18~\improve{+1.12} & 0.9651~\improve{+0.0100} \\
            NAFNet~\cite{chen2022nafnet} & 26.47 & 0.9475 & 29.28~\improve{+2.81} & 0.9604~\improve{+0.0129} \\
            DehazeFormer~\cite{song2023dehazeformer} & 27.02 & 0.9552 & 28.71~\improve{+1.69} & 0.9557~\improve{+0.0005}\\
            MS-DehazeNet & 28.18 & 0.9632 & 29.90~\improve{+1.72} & 0.9709~\improve{+0.0077} \\
            \bottomrule[0.15em]
    
    \end{tabular}}
    \label{table:ab_dehazer}
    \end{minipage}
    \begin{minipage}[c]{0.39\textwidth}
    \caption{Effect of $\Delta\beta$.}
    \vspace{0.5em}
    \centering
    \setlength{\tabcolsep}{3pt}
    \scalebox{0.9}{
    \begin{tabular}{l | c c  }
        \toprule[0.15em]
            $\Delta \beta$  & PSNR$\textcolor{black}{\uparrow}$ & SSIM$\textcolor{black}{\uparrow}$  \\
            \midrule[0.15em]
            w/o Rehazy & 28.18 & 0.9632  \\
            \midrule
            ${\rm U}(0.1, 0.25)$ & 29.43 & 0.9703  \\
            ${\rm U}(0.1, 0.5)$ & \best{29.90} & \best{0.9709} \\
            ${\rm U}(0.1, 0.75)$ & 29.76 & 0.9676  \\
            ${\rm U}(0.1, 1.0)$ & 29.31 & 0.9646\\
            \bottomrule[0.15em]
    \end{tabular}}
    
    \label{table:ab_beta}
        
    \end{minipage}
    \vspace{-0.5em}
\end{table*}

\vspace{-0.25em}
\subsection{Ablation Studies} \label{sec:ablation}

\paragraph{The effect of rehazy strategy.}

\begin{figure}
    \centering
    \includegraphics[width=1.0\linewidth]{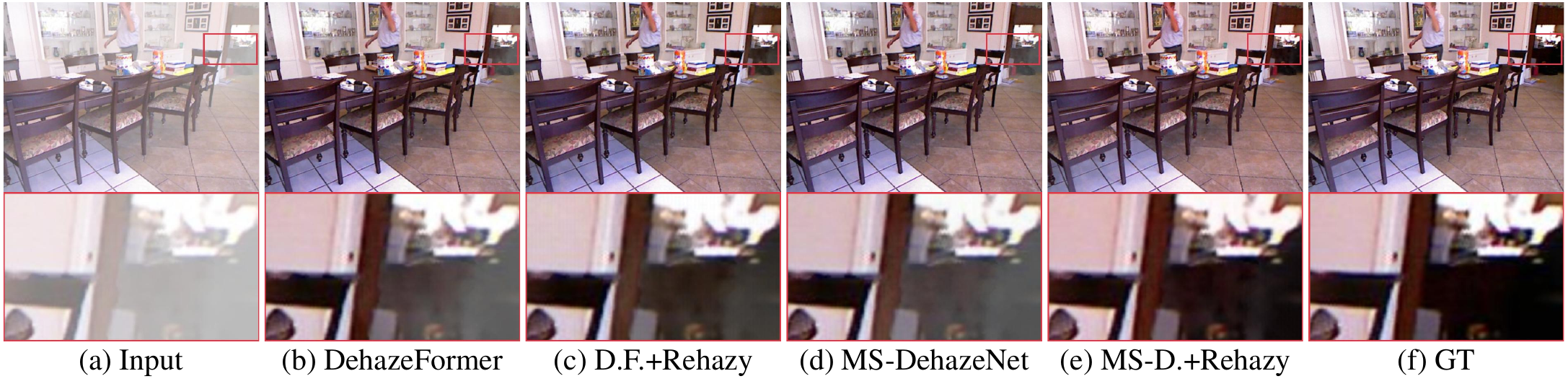}
    \vspace{-1.5em}
    \caption{
    Visual results of ablation study on rehazy strategy. ``D.F.'' and ``MS-D.'' denote the DehazeFormer and MS-DehazeNet, respectively. ``+Rehazy'' indicates adopting the rehazy strategy.
    }
    \vspace{-0.25em}
    
    \label{fig:ablation_dehazer_vis}
\end{figure}

To validate the effectiveness of the proposed rehazy strategy, we conduct an ablation study on the SOTS-Indoor dataset. Specifically, we compare dehazing models trained without and with the rehazy strategy, denoted as ``w/o Rehazy'' and ``w/ Rehazy''. Note that when the rehazy strategy is removed, our framework in Fig.~\ref{fig:framework} simplifies to the clean branch only. Furthermore, for a comprehensive evaluation, we employ four different restoration networks as the dehazer: U-Net~\cite{ronneberger2015unet}, NAFNet~\cite{chen2022nafnet}, DehazeFormer~\cite{song2023dehazeformer}, and MS-DehazeNet.
From the results presented in Tab.~\ref{table:ab_dehazer} and Fig.~\ref{fig:ablation_dehazer_vis}, we can draw two findings.
First, regardless of the network employed as the dehazer, the proposed rehazy strategy consistently benefits the dehazing task. With MS-DehazeNet, the rehazy strategy achieves a marginal improvement by 1.72 dB PSNR and 0.0077 SSIM. Moreover, when implementing the NAFNet, the improvement increases to 2.81 dB PSNR and 0.0129 SSIM. Additionally, as shown in Fig.~\ref{fig:ablation_dehazer_vis}, the rehazy strategy benefits more clearer haze removal. These improvements indicate that the rehazy strategy is capable of effectively capturing the haze features, thereby facilitating haze removal.
Second, compared to other networks, our MS-DehazeNet produces advanced performance, with at least a 0.62 dB and 0.0105 improvement in PSNR and SSIM, respectively, which convincingly demonstrates the effectiveness of MS-DehazeNet.

\vspace{-0.5em}
\paragraph{The effect of $\Delta \beta$ distribution.}

$\Delta \beta$ is a hand-crafted parameter used for rehazy image generation, which is randomly sampled from a pre-defined uniform distribution ${\rm U}(\Delta \beta_{min}, \Delta \beta_{max})$. To evaluate the effect of the distribution setting, we perform an ablation study on the SOTS-Indoor dataset. Specifically, the pre-defined distribution of $\Delta \beta$ is set to ${\rm U}(0.1, 0.25)$, ${\rm U}(0.1,0.5)$, ${\rm U}(0.1,0.75)$, and ${\rm U}(0.1,1.0)$, respectively. The experimental results are summarized in Tab.~\ref{table:ab_beta}.
It can be observed that across these ranges, the dehazing performance varies between 29.10 dB and 29.90 dB in PSNR, with the best result achieved when the range is set to ${\rm U}(0.1, 0.5)$.
Moreover, regardless of the pre-defined range, the proposed rehazy strategy consistently achieves improvements over the baseline without rehazy strategy.
Therefore, these findings suggest that even without careful tuning, a simple setting for $\Delta \beta$ distribution effectively benefits unpaired dehazing, underscoring the robustness of our approach.

\begin{figure}
    \centering
    \includegraphics[width=1.0\linewidth]{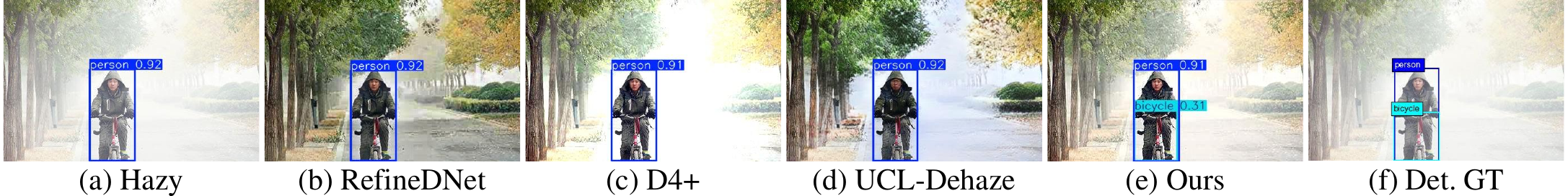}
    \vspace{-1.5em}
    \caption{
    Visual comparison of object detection on the RTTS dataset. {\color{blue} Blue} box and {\color{cyan} cyan} box represent the label of ``person'' and ``bicycle'', respectively.
    }
    \vspace{-0.75em}
    \label{fig:det_vis}
\end{figure}

\vspace{-0.5em}
\paragraph{Downstream task evaluation.}
Image degradations severely impact the high-level vision tasks, \eg, objection detection~\cite{yang2020advancing, fang2024real}.
To access the benefits of dehazing methods for such tasks, we apply a pre-trained detection model, YOLOv10~\cite{wang2024yolov10}, on the dehazed outputs of RTTS dataset, and evaluate the detection results using the mean Average Precision (mAP) metric based on the provided annotations. As presented in Tab.~\ref{table:rtts_results}, our method yields the best detection results on the dehazed images. Additionally, in the example illustrated in Fig.~\ref{fig:det_vis}, our approach successfully detects the bicycle, whereas other methods fail to do so.
As a result, these findings highlight the practical advantage of our method in facilitating high-level vision understanding.

%% file: Sec/05_Conclusion.tex
\vspace{-0.25em}
\section{Limitation} \label{sec:limitation}
\vspace{-0.25em}

Though effective, the proposed rehazy strategy is dependent on the ASM model, which may become unreliable in challenging scenarios with non-homogeneous illumination. This limitation constrains the proposed method’s ability to handle diverse haze degradations.
Additionally, our physics-based rehazy generation pipeline can only produce rehazy images with an airlight identical to that of the input hazy image, thereby restricting the diversity of the generated rehazy images.
Besides, although MS-DehazeNet achieves strong performance and competitive efficiency, it incurs higher computational overhead compared to D4+~\cite{yang2024d4}, limiting real-time application.
To overcome these limitations, it is suggested that more effective haze degradation models, generation pipelines, and dehazing networks be further explored.

\vspace{-0.25em}
\section{Conclusion}
\vspace{-0.25em}

In this paper, we propose an effective training strategy to facilitate unpaired image dehazing. This strategy explores the consistency of the underlying clean images across hazy images, and constructs hazy-rehazy pairs to facilitate effective learning of the real haze characteristics.
To obtain hazy-rehazy pairs, we develop a physics-based rehazy generation pipeline, where the ASM model and theoretical proof ensure the production of favorable rehazy images.
Furthermore, we introduce a dual-branch framework to train dehazing networks by leveraging this strategy, and design a multi-scale dehazing network to enhance dehazing capability and efficiency.
Extensive experiments demonstrate the superior dehazing performance of our method in both synthetic and real scenarios.

%% file: Sec/appendix.tex
\newpage
\appendix

\section*{Appendix}

\section{Derivation Detail}

In this section, we present the derivation details of Eq.~\ref{eq:rehazy_obj} and Eq.~\ref{eq:Ir2I0}, which are not included in our main paper due to space limitations.

\subsection{Derivation Detail of Eq.~\ref{eq:rehazy_obj}}

If plenty of hazy-clean pairs $\{(\mathbf{I},\mathbf{J})\}$ can be obtained, the optimization object of training a dehazing network in a supervised manner can be described as Eq.~\ref{eq:pair_obj}, as shown in the following:
\begin{equation} \label{eq:pair_obj_suppl}
\mathop{\rm argmin~}\limits_{\theta} \mathbb{E}_{(\mathbf{I},\mathbf{J})}\{\mathcal{L}(g_{\theta}(\mathbf{I}),\mathbf{J})\},
\end{equation}
where $\mathcal{L}$ denotes an empirical loss function. This indicates that the dehazed results should be close to the clean counterpart.
Assuming that a set of hazy images  $\{\mathbf{I}_i\}(i=0,1,...,n)$ maintain clean consistency and shared the clean counterpart $\mathbf{J}_0$, the optimization objective can become to:
\begin{equation}
\mathop{\rm argmin~}\limits_{\theta} \mathbb{E}_{\{\mathbf{J}_0\}} \mathbb{E}_{\{\mathbf{I}_i\}}\{\mathcal{L}(g_{\theta}(\mathbf{I}_i),\mathbf{J}_0)\}.
\end{equation}
Under the clean consistency assumption, an alternative approach can be adopted to achieve the same objective. Specifically, the dehazed output of an input hazy image $\mathbf{I}_0$ is encouraged to approximate its clean counterpart $\mathbf{J}_0$, while the dehazed results of the remaining hazy images $\{\mathbf{I}_i\}(i \neq 0)$ are constrained to align with that of $\mathbf{I}_0$, thereby implicitly aligning with $\mathbf{J}_0$ as well. Therefore, the optimization objective is equivalent to Eq.~\ref{eq:rehazy_obj}:
\begin{equation} \label{eq:rehazy_obj_suppl}
    \mathop{\rm argmin}\limits_{\theta}~\mathbb{E}_{\{\mathbf{J}_0\}}~[~
    \mathop{\underline{
    \mathbb{E}_{(i\neq0)}\{\mathcal{L}(g_{\theta}(\mathbf{I}_i),g_{\theta}(\mathbf{I}_0))\}
    }}\limits_{\mathcal{L}_{const}}
    +\mathop{\underline{
    \mathcal{L}(g_{\theta}(\mathbf{I}_0),\mathbf{J}_0)
    }}\limits_{\mathcal{L}_{dehazing}}
    ~].
\end{equation}

\subsection{Derivation Detail of Eq.~\ref{eq:Ir2I0}}

Eq.~\ref{eq:Ir2I0} describes the approach of the rehazy image generation based on the ASM model. The derivation details are provided below.

\begin{equation}
\begin{aligned}
    \mathbf{I}_r&=\mathbf{J}_r \mathbf{t}_r+A_r(1-\mathbf{t}_r)\\
    &=\mathbf{J}_0 \mathbf{t}_0 \Delta \mathbf{t}+(A_0+\Delta A)(1-\mathbf{t}_0\Delta \mathbf{t}) \\
    &=\mathbf{J}_0\mathbf{t}_0\Delta \mathbf{t}+A_0(1-\mathbf{t}_0\Delta \mathbf{t})+\Delta A(1-\mathbf{t}_0\Delta \mathbf{t}) \\  
    &=\mathbf{J}_0\mathbf{t}_0\Delta \mathbf{t}+A_0(1-\mathbf{t}_0\Delta \mathbf{t})+{\color{blue}A_0\Delta \mathbf{t} - A_0\Delta \mathbf{t}}+\Delta A(1-\mathbf{t}_0\Delta \mathbf{t}) \\    
    &=\mathbf{J}_0 \mathbf{t}_0 \Delta \mathbf{t} +A_0({\color{blue}\Delta \mathbf{t}}-\mathbf{t}_0\Delta \mathbf{t}) + A_0 -{\color{blue} A_0 \Delta \mathbf{t}} +\Delta A(1-\mathbf{t}_0\Delta \mathbf{t}) \\
    &=\mathbf{J}_0 \mathbf{t}_0 \Delta \mathbf{t} +A_0 \Delta \mathbf{t} (1-\mathbf{t}_0) + A_0 (1-\Delta \mathbf{t}) +\Delta A(1-\mathbf{t}_0\Delta \mathbf{t}) \\
    &=\Delta \mathbf{t}(\mathbf{J}_0 \mathbf{t}_0 +A_0(1-\mathbf{t}_0)) + A_0 (1-\Delta \mathbf{t}) +\Delta A(1-\mathbf{t}_0\Delta \mathbf{t}) \\
    &=\Delta \mathbf{t} \mathbf{I}_0 + A_0(1-\Delta \mathbf{t}) +\Delta A(1-\mathbf{t}_0\Delta \mathbf{t})
\end{aligned}.
\end{equation}

\section{Discussions}

\subsection{Discussion on Rehazy Strategy}

The critical challenge in unpaired image dehazing lies in effectively capturing real haze characteristics in the absence of hazy-clean image pairs.
To eliminate this challenge, existing methods adopt CycleGAN-like frameworks to generate synthetic hazy-clean pairs or utilize CUT-like frameworks to learn from unaligned hazy-clean patch pairs.
Specifically, CycleGAN-like methods~\cite{engin2018cycledehaze, chen2022unpaired, yang2024d4, lan2025exploiting} construct hazy-clean-hazy and clean-hazy-clean cycles to jointly learn haze removal and realistic haze synthesis. The core idea is to leverage adversarial loss to synthesize realistic hazy images from clean inputs, thereby producing plenty of hazy-clean image pairs for training networks in a supervised fashion.
In contrast, CUT-like methods~\cite{chen2022unpaired, wang2024ucl, wang2024odcr, luo2025farewell} typically sample patches from the hazy input, clean input, and dehazed output to form query-positive-negative pairs. These are incorporated into a contrastive learning framework to guide the network in learning haze removal.
Despite their progress, CycleGAN-based methods require careful design of both the generator and discriminator, while contrastive learning approaches depend heavily on the construction of effective query-positive-negative pairs. As a result, these two frameworks often exhibit training instability, which limits their dehazing performance.

To overcome the aforementioned limitations, we propose the rehazy strategy that leverages clean consistency, as introduced in Sec.~\ref{sec:intro}  of the main paper. Unlike previous approaches that focus on constructing hazy-clean or query-positive-negative pairs, our method forms realistic hazy-rehazy pairs to better capture real haze features.
This strategy benefits the unpaired dehazing task in two key aspects.
First, we adopt a synthetic approach to generate hazy images from clean inputs, enabling the network to acquire a fundamental dehazing capability through supervised training, which in turn stabilizes the optimization process.
Second, to enhance generalization to real-world scenarios, the rehazy strategy constructs effective training pairs by combining real hazy inputs with high-quality rehazy images, thereby facilitating the learning from real hazy images and promoting training stability.
As a result, the proposed rehazy strategy not only substantially improves dehazing performance in real-world scenarios but also provides an effective and generalizable training paradigm.
As introduced in Sec.~\ref{sec:ablation}, the proposed rehazy strategy is compatible with a wide range of dehazing networks and consistently enhances their performance.

\begin{figure}
    \centering
    \includegraphics[width=1.0\linewidth]{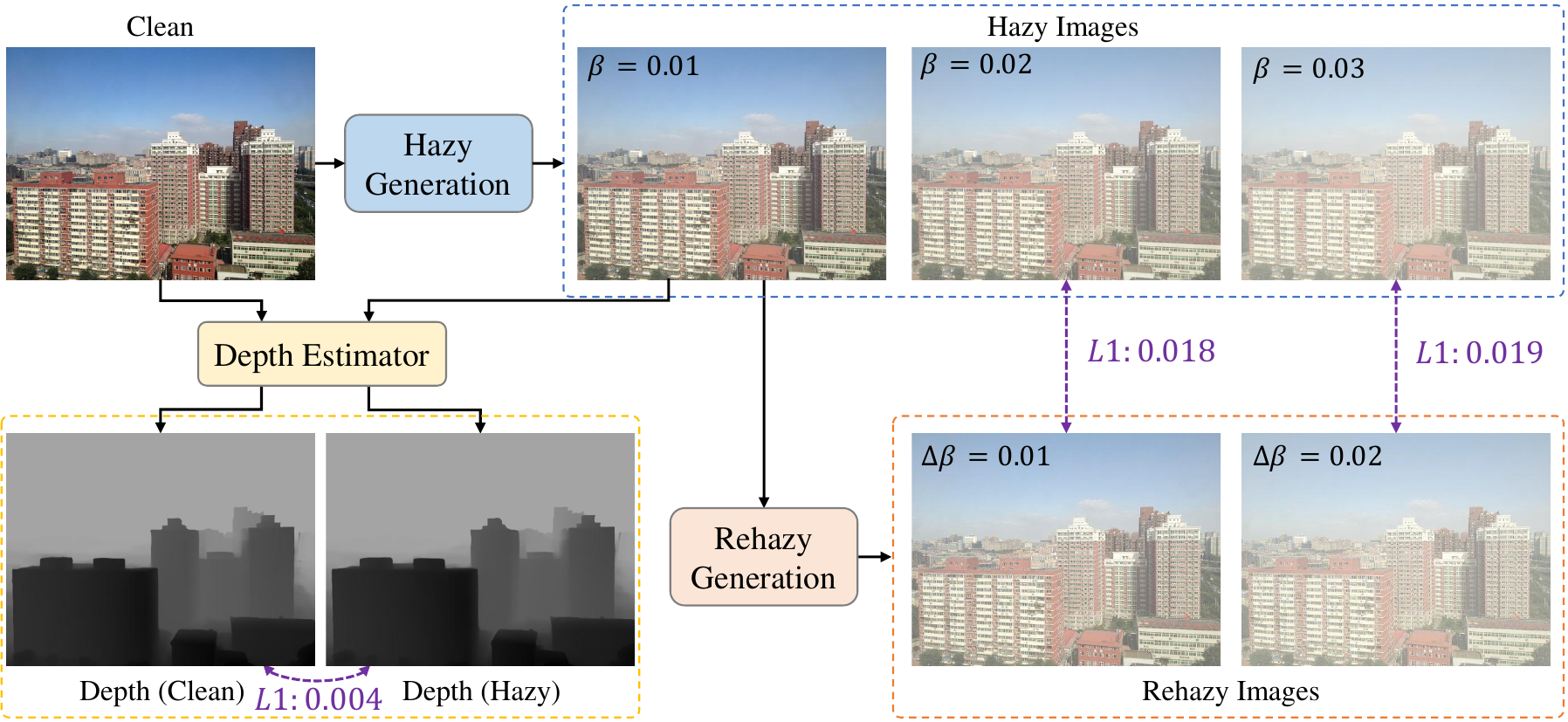}
    \caption{An ablation example illustrating the effectiveness of our physics-based rehazy generation.}
    \vspace{-1em}
    \label{fig:rehazy_proof}
\end{figure}

\subsection{Discussion on Physics-based Rehazy Generation}

As outlined in Sec.\ref{sec:rehazy} of the main paper, we develop a physics-based rehazy generation pipeline to construct hazy-rehazy image pairs. Specifically, grounded on the ASM model, we formulate a simple yet effective method requiring only three parameters: $A_0$, $\Delta \beta$, and $\mathbf{d_0}$, as illustrated in Eq.~\ref{eq:rehazy}, which are estimated through three practical strategies. A key approximation in these strategies is that, due to the strong robustness of the Depth Anything model, the depth estimated from the hazy image approximates that of the clean scene~\cite{ye2025prompthaze}.

To verify the correctness of the proposed pipeline, we conduct an ablation experiment, as shown in Fig.~\ref{fig:rehazy_proof}.
Specifically, starting from a clean image, we adopt the ASM model to generate hazy images with scattering coefficients $\beta =0.01$, $\beta =0.02$, and $\beta =0.03$, respectively.
We then employ a pre-trained Depth Anything model to estimate the depth of the clean image and the hazy image with $\beta =0.01$. The L1 divergence between the two depth maps is found to be only 0.004, suggesting that the estimated depth from the hazy image serves as a reliable approximation of the clean image’s depth.
Moreover, although prior work~\cite{ye2025prompthaze} has noted that the Depth Anything model may introduce errors in few severe haze conditions, we argue that this approximation remains effective in our framework.
This is because the regions with thick haze exhibit pixel values close to the airlight. According to Eq.~\ref{eq:rehazy}, even with an imperfect transmission map during the rehazy process, introducing additional haze to such areas consistently yields intensities near the airlight, thereby still producing visual plausible rehazy results.

Additionally, using the hazy image with $\beta = 0.01$ as input, we apply the proposed rehazy generation pipeline to produce rehazy images with $\Delta \beta = 0.01$ and $\Delta \beta = 0.02$. According to the Eq.~\ref{eq:rehazy}, we have $\beta_r = \beta_0 + \Delta \beta$, where $\beta_r$ and $\beta_0$ are the scattering coefficients of generated rehazy images and hazy inputs, respectively. This indicates that the rehazy image with $\Delta \beta = 0.01 $ and $\Delta \beta = 0.02 $ should closely resemble those hazy images previously synthesized with $\beta =0.02$ and $\beta =0.03$, respectively. Hence, we further compute the L1 difference between them, yielding minimal values of 0.018 and 0.019. These results indicate that the generated rehazy images exhibit strong similarity to the desired hazy images, confirming the correctness of our rehazy generation pipeline.

\section{More Ablation Studies}

\subsection{Ablation Study on the Number of Generated Rehazy Images}

In the hazy branch, we utilize the physics-based rehazy generation pipeline to synthesize rehazy images from a given hazy input. To evaluate the impact of the number of generated rehazy images, we conduct an ablation study by varying the number from 0 to 5 on the SOTS-Indoor dataset. Note that the value of 0 indicates stripping the rehazy strategy and only adopting the clean branch. As presented in Tab.~\ref{table:ab_rehazynum}, it can be observed that dehazing performance shows minimal variation when the number of rehazy images ranges from 1 to 5. For example, PSNR ranges only from 29.90 dB to 30.02 dB. This is because, similar to the effect of batch size, generating multiple rehazy images in a single iteration is approximately equivalent to generating a single rehazy image across multiple iterations. To lower GPU memory demands, we adopt the setting of generating one rehazy image per training iteration in our experiments.

\begin{table*}[h]
    \centering
    \vspace{-1em}
    \caption{Ablation study on the number of generated rehazy images.}
    \vspace{0.5em}
    \setlength{\tabcolsep}{4.5pt}
    \scalebox{0.9}{
    \begin{tabular}{c | c c c c c c }
            \toprule[0.15em]
            \#Rehazy images & 0 & 1 & 2 & 3 & 4 & 5 \\
            
            \midrule[0.15em]
            PSNR$\textcolor{black}{\uparrow}$ & 28.18 & 29.90 & 29.95 & 30.02 & 29.91 & 29.90  \\
            SSIM$\textcolor{black}{\uparrow}$ & 0.9632 & 0.9709 & 0.9677 & 0.9710 & 0.9692 & 0.9690 \\
          
            \bottomrule[0.15em]
    \end{tabular}}
    \label{table:ab_rehazynum}
    \vspace{-0.75em}
\end{table*}

\subsection{Ablation Study on the Scale of MS-DehazeNet}

Given that the proposed MS-DehazeNet adopts a multi-scale architecture, we further investigate its efficiency and performance under varying scales. Specifically, we configure the network with 2, 3, and 4 scales, respectively, and conduct evaluations on the SOTS-Indoor dataset. As shown in Tab.~\ref{table:ab_msscale}, dehazing performance consistently improves with increasing network depth, demonstrating the effectiveness of the progressive learning strategy from lower to higher scales. However, increasing the scale from 3 to 4 results in a 4.58G MACs increase (approximately $40\%$) while yielding only a 0.25 dB improvement in PSNR. Therefore, for a trade-off between computational efficiency and dehazing performance, we adopt a scale depth of 3 in our method.

\begin{table*}[h]
    \centering
    \setlength{\tabcolsep}{4.5pt}
    \vspace{-1em}
    \caption{Ablation study on the scales of MS-DehazeNet.}
    \vspace{0.5em}
    \scalebox{0.9}{
    \begin{tabular}{c | c c | c c }
            \toprule[0.15em]
            \#Scale & Params(M) & MACs(G) & PSNR$\textcolor{black}{\uparrow}$ & SSIM$\textcolor{black}{\uparrow}$  \\
            
            \midrule[0.15em]
             2 & 0.88  & 7.39 & 28.76 & 0.9620  \\
             3 & 1.57 & 11.09 & 29.90 & 0.9709  \\
             4 & 4.21 & 15.67 & 30.15 & 0.9710 \\
          
            \bottomrule[0.15em]
    \end{tabular}}
    \label{table:ab_msscale}
    \vspace{-0.75em}
\end{table*}

\subsection{Ablation Study on Loss Functions}

In the hazy branch of our framework, we introduce two losses, $\mathcal{L}_{depth}$ and $\mathcal{L}_{clip}$, to enforce invariance between hazy and dehazed images. To assess their effect, we perform an ablation study on the SOTS-Indoor dataset by removing each loss individually, denoted as ``w/o $\mathcal{L}_{depth}$ '' and ``w/o $\mathcal{L}_{clip}$''. 
As presented in Tab.~\ref{table:ab_loss}, stripping $\mathcal{L}_{depth}$ or $\mathcal{L}_{clip}$ results in a PSNR drop of 0.39 dB and 0.38 dB, respectively, indicating that modeling these invariances facilitates effective learning of real-world haze.
In addition, we study the effectiveness of the coarse-to-fine design in the MS-DehazeNet by replacing Eq.~\ref{eq:ms_loss} with Eq.~\ref{eq:clean_loss} and Eq.~\ref{eq:const_loss}, shown as ``w/o Coarse-to-Fine'' in Tab.~\ref{table:ab_loss}. The resulting 0.29 dB PSNR reduction confirms the advantage of this progressive strategy in addressing such a low-frequency degradation. Moreover, the results of these ablation studies outperform that without the rehazy strategy (\ie, ``w/o Rehazy''), providing additional evidence for the advancement and superiority of our proposed rehazy strategy.

\begin{table*}
    \centering
    \vspace{-1em}
    \caption{Ablation study on loss functions.}
    \vspace{0.5em}
    \scalebox{0.9}{
    \begin{tabular}{l | c | c c c c }
        \toprule[0.15em]
            \multirow{2}{*}{Metrics} & \multirow{2}{*}{w/o Rehazy} & \multicolumn{4}{c}{w/ Rehazy}  \\
            & & w/o $\mathcal{L}_{depth}$ & w/o $\mathcal{L}_{clip}$ & w/o Coarse-to-Fine & Ours \\
        \midrule[0.15em]
            PSNR$\textcolor{black}{\uparrow}$ & 28.18 & 29.51 & 29.52 & 29.61 & 29.90 \\
            SSIM$\textcolor{black}{\uparrow}$ & 0.9632 & 0.9660 & 0.9694 & 0.9639 & 0.9709 \\
        \bottomrule[0.15em]

    \end{tabular} 
    }\label{table:ab_loss}
\end{table*}

\section{More Visual Results}

\begin{figure}
    \centering
    \includegraphics[width=1.0\linewidth]{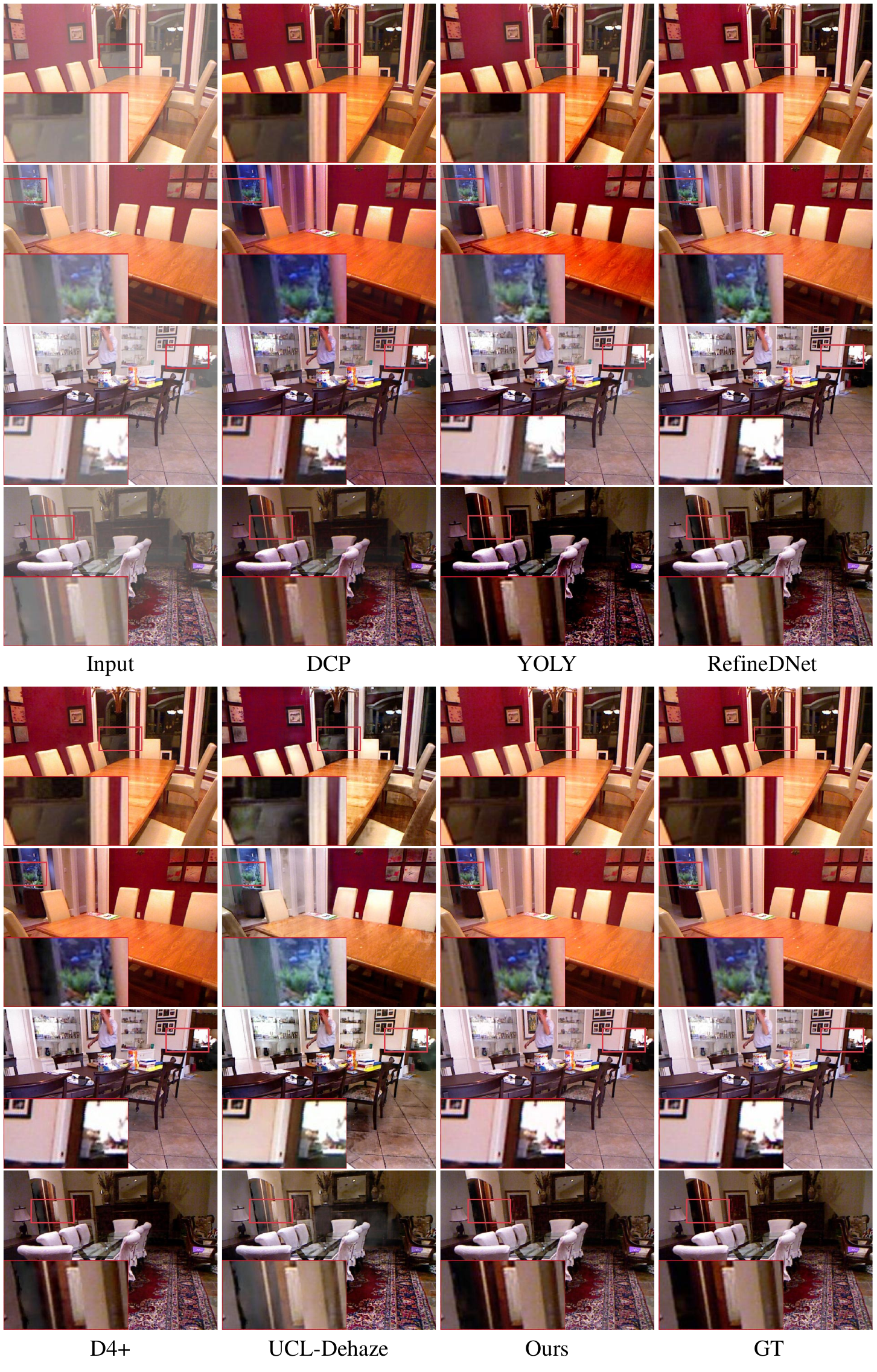}
    \caption{
    Visual comparison on the SOTS-Indoor dataset.
    }
    \label{fig:suppl_its}
\end{figure}

\begin{figure}
    \centering
    \includegraphics[width=1.0\linewidth]{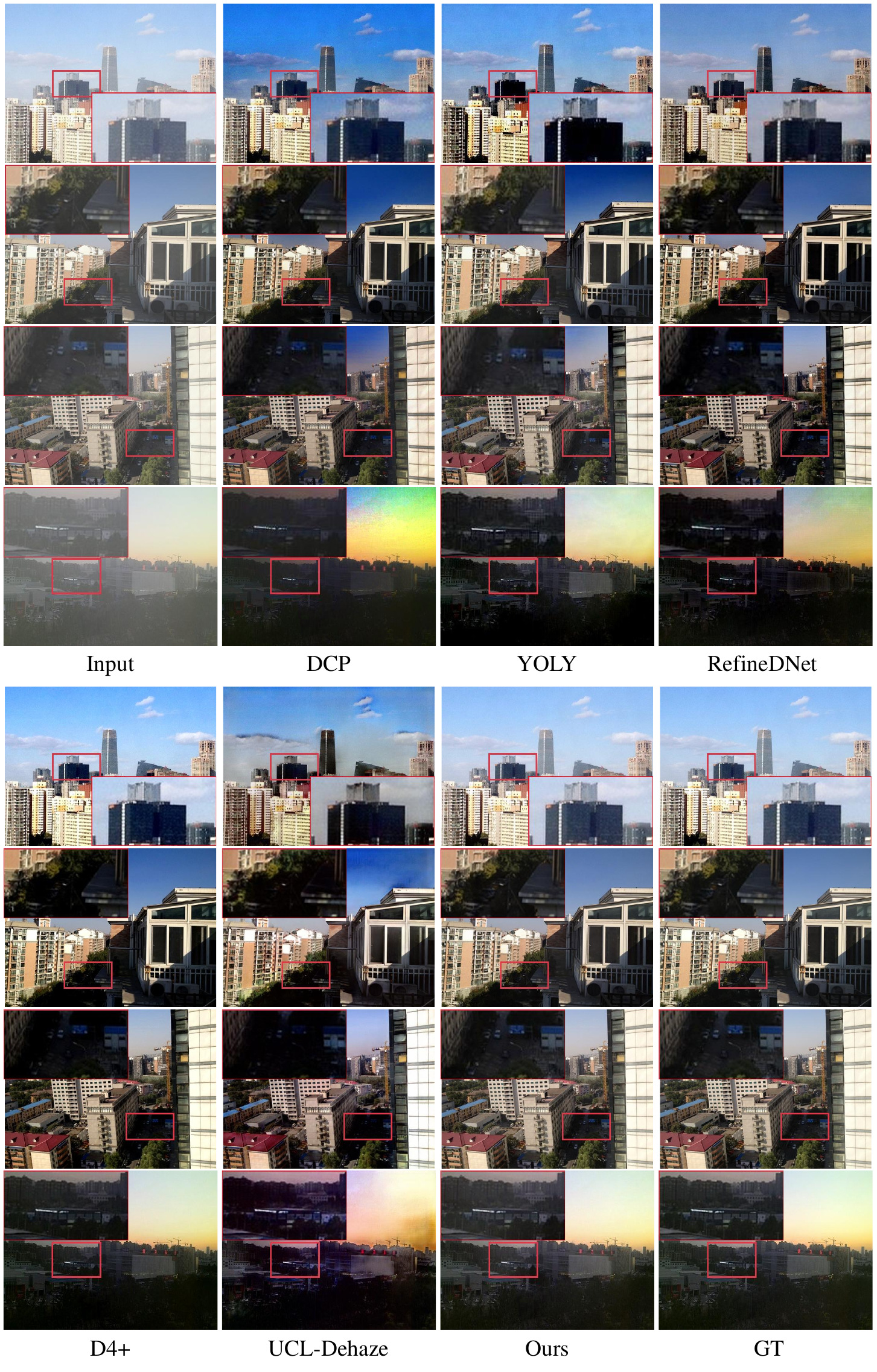}
    \caption{
    Visual comparison on the SOTS-Outdoor dataset.
    }
    \label{fig:suppl_ots}
\end{figure}

\begin{figure}
    \centering
    \includegraphics[width=1.0\linewidth]{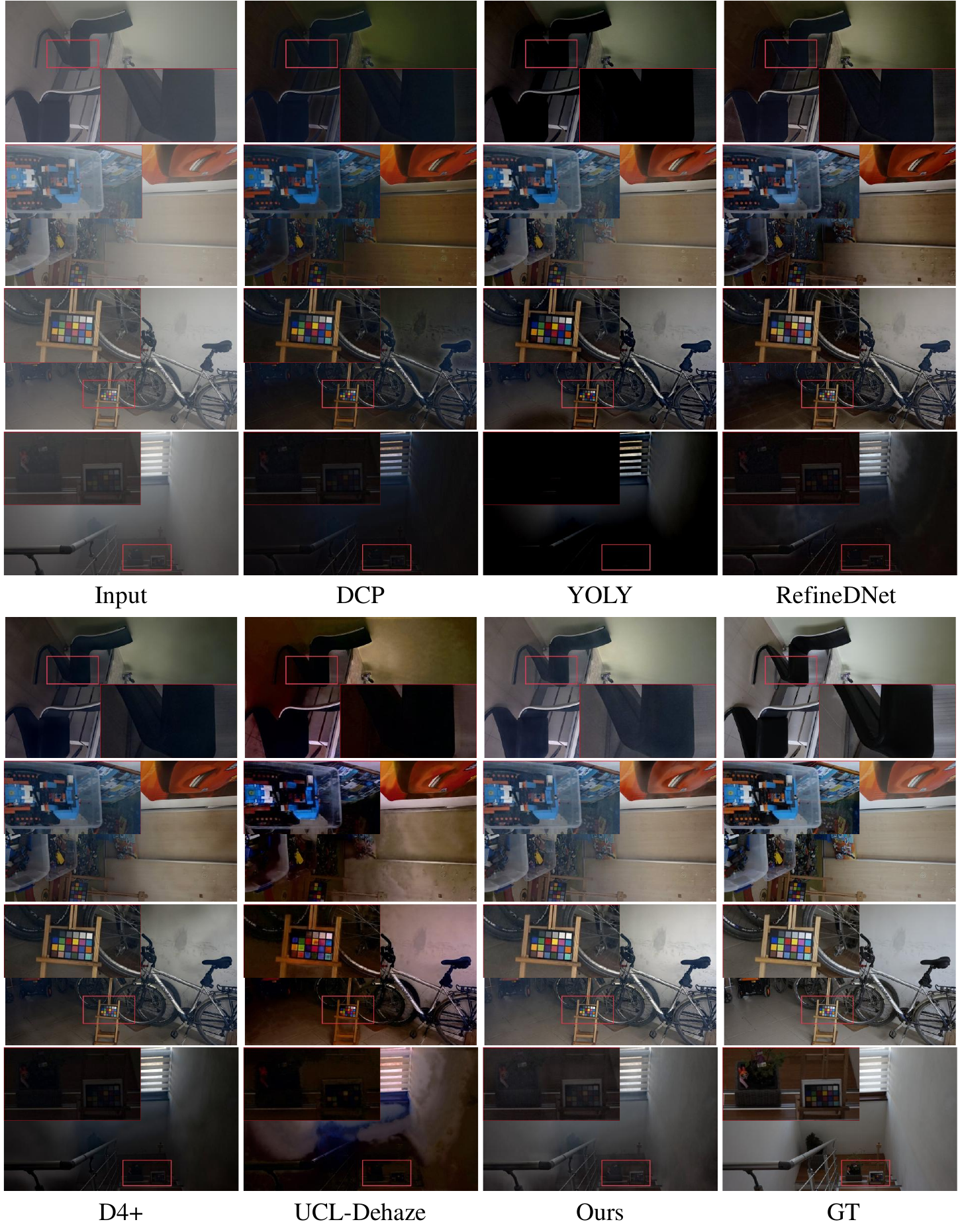}
    \caption{
    Visual comparison on the I-HAZE dataset.
    }
    \label{fig:suppl_ihaze}
\end{figure}

\begin{figure}
    \centering
    \includegraphics[width=1.0\linewidth]{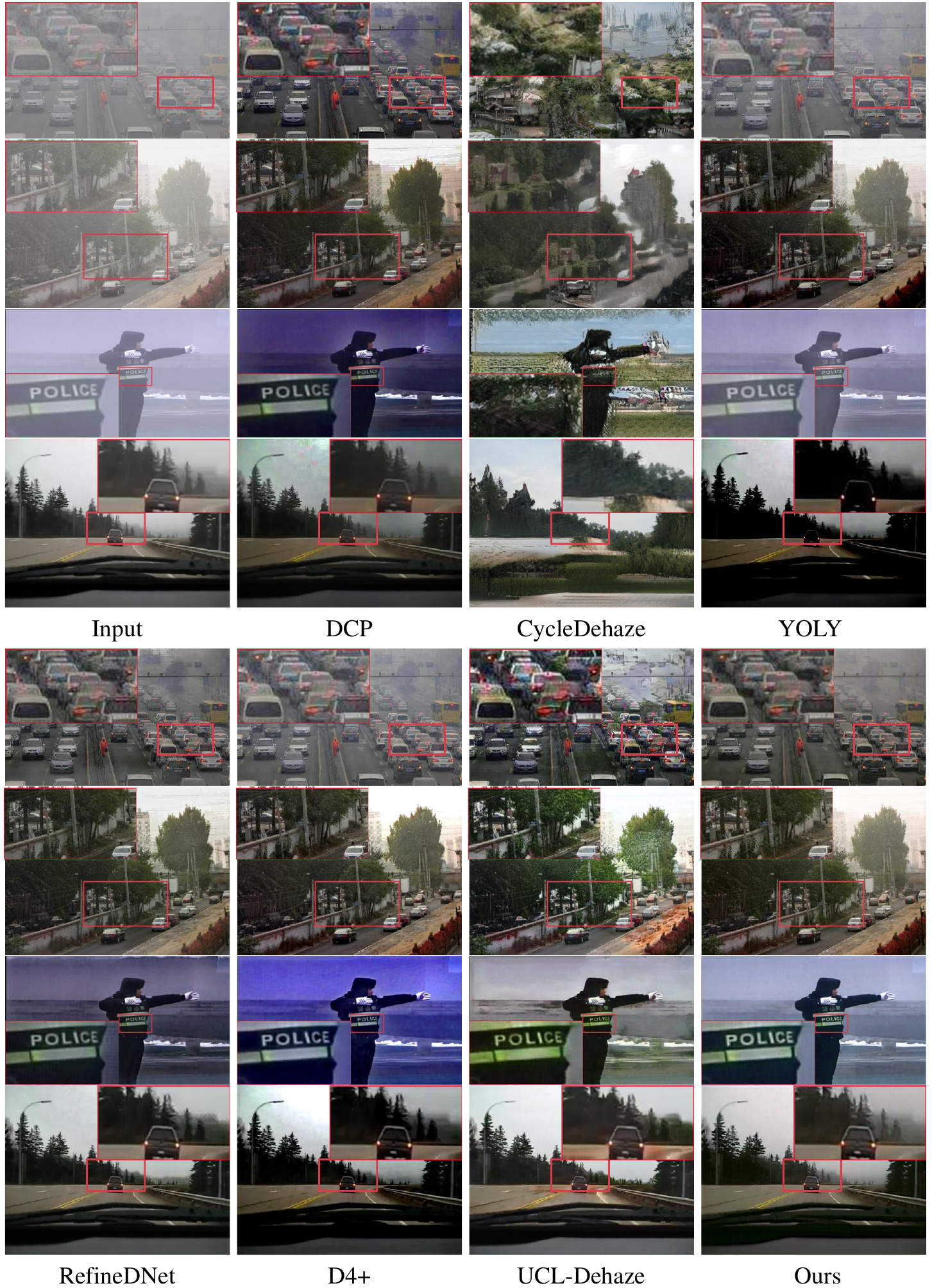}
    \caption{
    Visual comparison on the RTTS dataset.
    }
    \label{fig:suppl_rtts}
\end{figure}

We present more visual comparisons on the SOTS-Indoor, SOTS-Outdoor, I-HAZE, and RTTS datasets, as shown in Fig.~\ref{fig:suppl_its}, Fig.~\ref{fig:suppl_ots}, Fig.~\ref{fig:suppl_ihaze}, and Fig.~\ref{fig:suppl_rtts}, respectively.
In Fig.~\ref{fig:suppl_its}, our method demonstrates accurate estimation of haze density and effective color restoration. When generalized to the outdoor scenes in Fig.~\ref{fig:suppl_ots}, it consistently produces visually pleasing results, particularly in sky regions. In the more challenging scenarios of Fig.~\ref{fig:suppl_ihaze}, our approach maintains output naturalness, while other methods tend to introduce noticeable artifacts. For the real-world scenes in Fig.~\ref{fig:suppl_rtts}, our method continues to generate favorable and realistic results. These results further confirm the effectiveness of our method.

\section{Boarder Impact} \label{sec:boarder_impacts}
Unpaired image dehazing is a crucial task in image restoration, aimed at restoring clean scenes from haze-corrupted images in real-world scenarios.
In real-world applications, effective dehazing benefits downstream vision tasks, such as object detection, segmentation, and depth estimation, thereby supporting broader applications including autonomous driving and security monitoring.
To tackle the training instability issue of existing methods, we propose a novel training strategy for unpaired image dehazing. By developing a training framework and designing a dehazing network, we achieve state-of-the-art performance in dehazing tasks. Additionally, ablation studies show that our method is capable of facilitating downstream tasks. Notably, image dehazing techniques, including ours, have not exhibited any negative societal impacts.

